\documentclass[letterpaper, 10 pt, conference]{ieeeconf}  % Comment this line out if you need a4paper

\IEEEoverridecommandlockouts                              % This command is only needed if 
\usepackage{cite}
\usepackage{amsmath,amssymb,amsfonts}
\usepackage{algorithmic}
\usepackage{graphicx}
\usepackage{textcomp}
\usepackage{xcolor}
\usepackage{verbatim}
\usepackage{float}
\usepackage{multicol}
\usepackage{multirow}
\usepackage{gensymb}
\usepackage{subfigure}
\usepackage{array}
\usepackage{booktabs}
\usepackage{multirow}
\usepackage{amssymb}

\title{\LARGE \bf
JointLoc: A Real-time Visual Localization Framework for Planetary UAVs Based on Joint Relative and Absolute Pose Estimation
}

\author{Xubo Luo$^{1,2,3}$, Xue Wan$^{2,3}$, Yixing Gao$^{4}$, Yaolin Tian$^{1,2,3}$, Wei Zhang$^{2,3}$ and Leizheng Shu$^{2,3}$% <-this % stops a space
\thanks{*This work was supported by the National Natural Science Foundation of China (NSFC, No. 42171445).}% <-this % stops a space
\thanks{$^{1}$Xubo Luo and Yaolin Tian are with the School of Aeronautics and Astronautics,
        University of Chinese Academy of Sciences, Beijing 101408, China
        {\tt\small \{luoxubo23, tianyaolin21\}@mails.ucas.ac.cn}}%
\thanks{$^{2}$Xubo Luo, Yaolin Tian, Xue Wan, Wei Zhang, and Leizheng Shu are with the 
Technology and Engineering Center for Space Utilization, Chinese Academy of Sciences, Beijing 100094, China
        {\tt\small \{wanxue, zhangwei, shuleizheng\}@csu.ac.cn}}%
\thanks{$^{3}$Xubo Luo, Yaolin Tian, Xue Wan, Wei Zhang, and Leizheng Shu are with the 
Key Laboratory of Space Utilization, Chinese Academy of Sciences, Beijing 100094, China
        }%
\thanks{$^{4}$Yixing Gao is with the School of Artificial Intelligence, Jilin University, Changchun 130015, China
        {\tt\small gaoyixing@jlu.edu.cn}}%
}

\begin{document}

\maketitle
\thispagestyle{empty}
\pagestyle{empty}

%%%%%%%%%%%%%%%%%%%%%%%%%%%%%%%%%%%%%%%%%%%%%%%%%%%%%%%%%%%%%%%%%%%%%%%%%%%%%%%%
\begin{abstract}

Unmanned aerial vehicles (UAVs) visual localization in planetary aims to estimate the absolute pose of the UAV in the world coordinate system through satellite maps and images captured by on-board cameras. However, since planetary scenes often lack significant landmarks and there are modal differences between satellite maps and UAV images, the accuracy and real-time performance of UAV positioning will be reduced. In order to accurately determine the position of the UAV in a planetary scene in the absence of the global navigation satellite system (GNSS), this paper proposes JointLoc, which estimates the real-time UAV position in the world coordinate system by adaptively fusing the absolute 2-degree-of-freedom (2-DoF) pose and the relative 6-degree-of-freedom (6-DoF) pose. Extensive comparative experiments were conducted on a proposed planetary UAV image cross-modal localization dataset, which contains three types of typical Martian topography generated via a simulation engine as well as real Martian UAV images from the Ingenuity helicopter. JointLoc achieved a root-mean-square error of 0.237m in the trajectories of up to 1,000m, compared to 0.594m and 0.557m for ORB-SLAM2 and ORB-SLAM3 respectively. The source code will be available at https://github.com/LuoXubo/JointLoc.

\end{abstract}

%%%%%%%%%%%%%%%%%%%%%%%%%%%%%%%%%%%%%%%%%%%%%%%%%%%%%%%%%%%%%%%%%%%%%%%%%%%%%%%%
\section{INTRODUCTION}
Planetary exploration, vital for expanding human habitation beyond Earth, has seen a rise in interest in unmanned aerial vehicles (UAVs) due to their portability and exploration capabilities~\cite{drones6010004}. Notably, the Ingenuity helicopter of NASA deployed to Mars in September 2021 has completed 72 flights, capturing valuable imagery of the Martian surface~\cite{balaram2021ingenuity}. Accurate positioning of UAVs is crucial for mission success, as evidenced by a January 2024 incident where the Ingenuity experienced communication loss after a hard landing due to positioning challenges in a featureless area~\cite{FisherJohnson2024}, concluding its pioneering flight on Mars.

\begin{figure}[htp]
    \centering
    \includegraphics[width=1.0\linewidth]{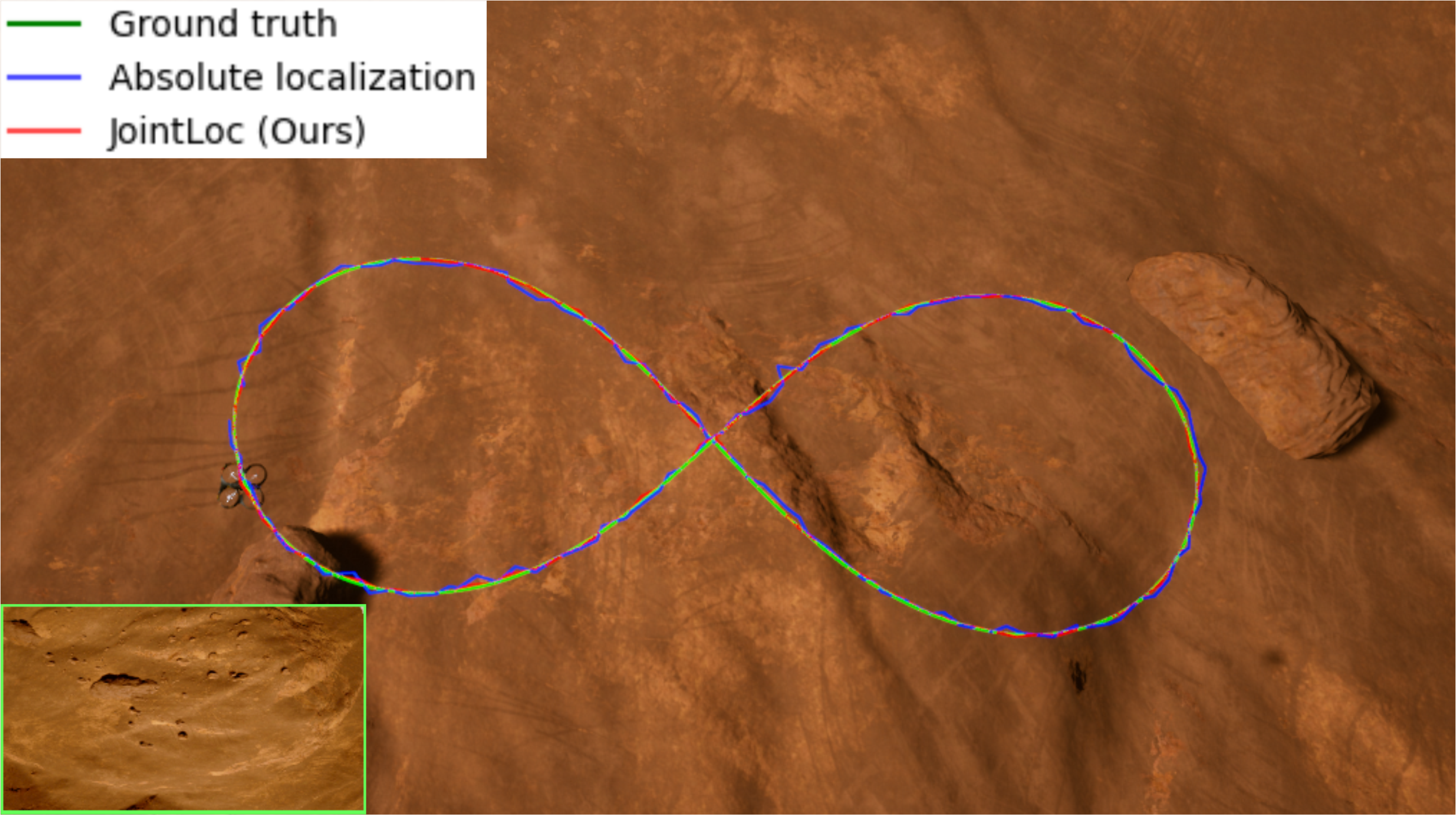}
    \caption{JointLoc accurately estimates the UAV poses. In contrast, the results of absolute localization are rough and pose a great challenge to the control algorithm. The results of relative localization are in the local coordinate system and require manual alignment to the world coordinate system.}
    \label{fig:comparison}
\end{figure}

In contrast to the highly developed navigation system, such as the Global Navigation Satellite System (GNSS), on Earth, autonomous UAV navigation in a planetary environment presents several challenges. Firstly, the absence of GNSS means that the geographic position of the UAV cannot be directly determined. Secondly, the planetary environment usually lacks salient targets and rich texture information which prevents some visual position estimation methods commonly used in Earth scenes from being directly applied in planetary exploration missions. In addition, the limited computing power of mobile computing platforms prevents the most popular methods from being used for positioning.

Mainstream localization methods in planetary exploration missions are typically categorized into absolute localization methods and relative localization methods~\cite{pavlenko2019wireless}. Absolute localization involves estimating the 2-DoF position of the vehicles within the world coordinate system by comparing visual data collected by onboard sensors with existing satellite remote sensing maps or terrain maps~\cite{5649051}. For instance, Geromichalos et al.~\cite{geromichalos2020slam} employed particle filters and scan matching with existing maps to reduce positioning errors for planetary rovers. It is evident that most current absolute localization methods are primarily utilized for planetary rovers at relatively low speeds. However, this presents a challenge for absolute localization, as large-scale map positioning necessitates considerable time and computational resources, rendering it unsuitable for high-speed moving UAVs. Furthermore, modal differences between satellite maps and UAV images, such as scale, viewing angle, and lighting, mean that absolute localization cannot guarantee the accuracy of the results.

Relative localization estimates the 6-DoF pose of the vehicles relative to the starting point. The main difference between relative and absolute localization is that the geographic position, such as latitude and longitude, can not be determined in the relative localization method. The most widely used method for relative localization in terrestrial scenes is the visual odometry (VO)~\cite{hong2021visual, giubilato2018experimental, giubilato2021gpgm}, which provides rapid positioning and high success rates. However, VO primarily estimates the local 6-DoF poses of the vehicles through inter-frame matching, rather than in the world coordinate system. Particularly in planetary environments like textureless Mars deserts, traditional operator-based VO struggles to extract a sufficient number of map points, leading to tracking lost~\cite{colosi2019better}. By fusing the 2-DoF absolute pose in the world coordinate system with the relative 6-DoF pose generated by VO, the 6-DoF pose in the world coordinate system can be estimated.

Although several studies have been carried out on either absolute or relative localization for planetary UAV, they do not take into account the consumption of computing resources by the absolute and relative poses fusion frequency~\cite{b1} and the global localization methods based on large-scale maps~\cite{pavlenko2019wireless}. Besides, in the case of textureless areas on a planetary surface, there is no effective method to deal with the deviation between absolute localization and relative localization. Therefore, this paper introduces JointLoc, a localization framework that estimates the 6-DoF pose of the planetary UAVs in world coordinate systems. JointLoc adaptively fuses absolute and relative poses, reducing the time consumption of absolute localization and also converting relative poses into the world coordinate system. A simple comparison of the UAV trajectories estimated by JointLoc and the absolute localization method is shown in Fig.~\ref{fig:comparison}.

The contributions of this paper can be summarized as follows:
\begin{itemize}
    \item A loosely coupled localization framework is proposed for planetary UAVs, which adaptively fuses relative 6-DoF and absolute 2-DoF poses, to estimate the 6-DoF poses of the UAV within the world coordinate system.
    \item For UAV localization in large-scale satellite maps, a coarse-to-fine absolute localization module is proposed to reduce time consumption. The deep features are used to increase the robustness of the module in textureless environments with few landmarks.
    \item In order to enable JointLoc to accurately estimate the pose even when the absolute and relative pose results are inconsistent, an adaptive confidence mechanism is designed for the absolute pose so that the pose fusion module can adjust the fusion weights of different poses based on the confidence.
    \item A planetary UAV dataset based on simulation and real-shot data is proposed. The dataset includes 6 sets of trajectories of three types of terrain: gravels, mountains, and craters, with a total of 3,000 images from a UAV perspective. At the same time, manual calibration generated 105 real-shot images and 95 simulated images of the Ingenuity.
\end{itemize}

\section{Related Works}
\subsection{Image matching based absolute localization}
Depending on the method of image matching, image-based absolute localization can be categorized into template-based and feature-based approaches. Template-based absolute localization has limitations, which uses the UAV image as a template and slides it over a geo-referenced map to determine the current position of the UAV. It relies solely on pixel differences between the UAV and satellite images, neglecting style, scale, and resolution differences. Researchers have explored using Scale-Invariant Feature Transform~\cite{b4} (SIFT) descriptors to improve robustness. SIFT features are scale-invariant and remain consistent under various transformations, enabling accurate correspondence between UAV and satellite images. This allows for estimating the position within the remote sensing map based on SIFT features and homography estimation.

Subsequently, more feature descriptors for image matching were developed, such as SURF~\cite{bay2008speeded}, ORB~\cite{rublee2011orb}, and AKAZE~\cite{alcantarilla2011fast}, among others. However, these methods rely on pixel variations in images and often struggle to extract features effectively in regions with limited texture information, such as plains and deserts. Hence, some researchers explored deep features for robust inter-domain image matching. In 2020, Sarlin et al. enhanced matching accuracy with SuperGlue~\cite{sarlin2020superglue}, using local and global attention mechanisms on SuperPoint~\cite{detone2018superpoint} features. However, direct image matching struggled due to UAV-satellite scale differences. In 2022, Luo~\cite{10137193} proposed a scale-adaptive image-based localization method, focusing on global coarse and local precise localization to boost accuracy. These methods prioritize image matching but neglect 3D spatial data, limiting 6-DoF pose estimation. In addition, without utilizing the temporal relationship of the UAV image sequence, the trajectory estimated by the absolute localization method is not smooth, which poses a great challenge to the UAV control system.

\subsection{SLAM for UAV localization}
Simultaneous Localization and Mapping (SLAM) has garnered considerable attention as a technology capable of conducting real-time positioning and map construction, independent of external sensors such as the Global Positioning System (GPS). In 2017, Mur-Artal et. al proposed ORB-SLAM2~\cite{b14}, a monocular visual SLAM system, utilizing ORB features for robust mapping and localization with a single camera. Renowned for its real-time performance and versatility, it is a popular choice in computer vision and robotics for both indoor and outdoor applications. 
Researchers have proposed various SLAM algorithms for UAV localization, but they face limitations due to the absence of depth information, making it challenging to estimate the 6-DoF pose in the world coordinate system. This hinders the use of SLAM for UAV localization tasks. To address this, Wan et al.~\cite{b1} introduced a terrain-assisted SLAM system in 2022 that matches UAV images with terrain maps to obtain the world coordinates of the UAV. This system then combines absolute localization results with visual odometry from SLAM using the LBA algorithm, ultimately determining the UAV 6-DoF pose in the world coordinate system.

However, there are still some unresolved issues. The absolute localization algorithm in~\cite{b1} uses phase correlation which is unreliable in planetary imagery with large scale differences. In addition, it uses fixed weights when fusing absolute and relative poses, without considering that wrong absolute poses will affect positioning accuracy.

\section{Method}
\begin{figure*}[htb]
    \centering
    \includegraphics[width=\linewidth]{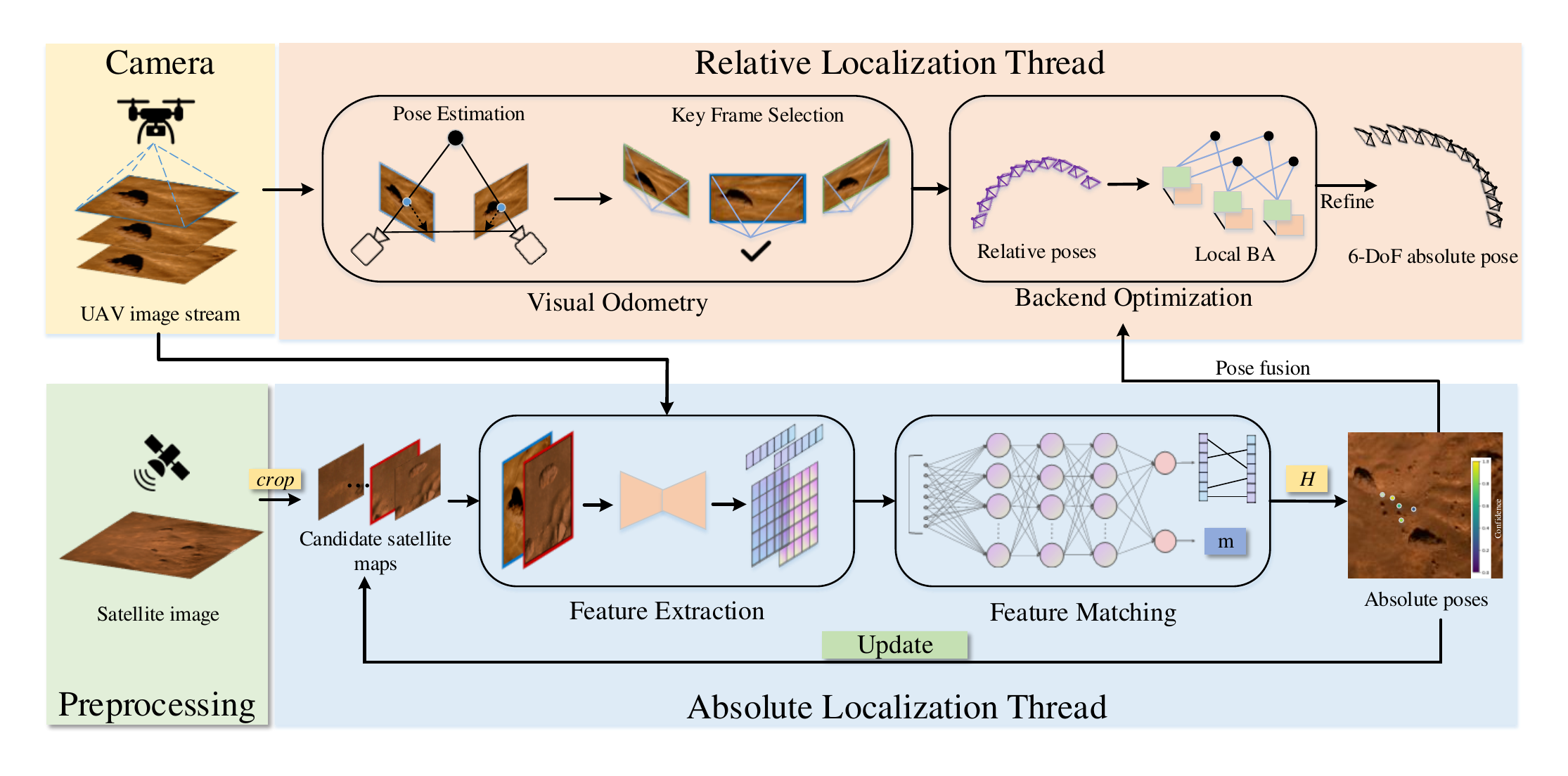}
    \caption{The pipeline of the proposed JointLoc consists of two threads: the absolute localization thread and the relative localization thread. In the relative localization thread, the UAV image is fed into visual odometry for relative pose estimation and keyframe selection. Simultaneously, the absolute localization module matches the UAV image with the candidate satellite map to estimate the absolute 2-DoF pose of the UAV. These absolute localization and visual odometry results are jointly processed by a pose fusion-based local bundle adjustment (LBA) algorithm. Then the transformation matrix from the relative coordinate system to the world coordinate system is calculated, resulting in the motion trajectory in the world coordinate system.}
    \label{fig:pipeline}
\end{figure*}

\subsection{Localization framework for planetary UAVs}
%JointLoc answers the following questions: 1) what is the optimal processing frequency for absolute and relative localization to ensure the safety of planetary exploration while saving energy as much as possible? 2) In the case of textureless areas on a planetary surface, how to deal with the deviation between absolute localization and relative localization?
In order to accurately estimate the 6-DoF pose of the UAV in real time, this paper proposes JointLoc based on joint relative and absolute pose estimation, as shown in Fig.~\ref{fig:pipeline}. Performing absolute localization for each frame of the UAV images will consume a lot of computing resources. Therefore, in order to ensure the real-time performance of JointLoc, this paper performs absolute localization at a fixed frequency, and the localization results are passed to the back-end module of relative localization for pose fusion. The details of pose fusion will be presented in Sec.\ref{fusion}.

When the absolute localization result is wrong, adding the absolute pose to the back-end module for pose fusion will reduce the positioning accuracy of JointLoc. In order to assign appropriate weights to different absolute localization results, a novel confidence mechanism based on global and local distribution characteristics for absolute localization results is proposed. Based on the confidence, soft pose fusion can still be performed when the absolute pose and relative pose deviations are large. The detailed design of the confidence mechanism can be seen in Sec.\ref{confidence}.

\subsection{Coarse-to-fine absolute localization}\label{AA}
To swiftly and accurately locate the given UAV image $\mathcal{I}_{\text{uav}}$ within the world coordinate system based on the satellite map $\mathcal{I}_{\text{sat}}$, we propose a coarse-to-fine absolute localization module. This module initially partitions the satellite map into several local maps. Utilizing the matching results, it anticipates the local map where the UAV may be positioned in the subsequent moment, thereby minimizing the need to re-search all local maps and consequently reducing the positioning time.

\subsubsection{Local satellite maps partitions}
To address the challenge posed by the large scale of the satellite map $\mathcal{I}_{\text{sat}}$ and its direct matching with the UAV image $\mathcal{I}_{\text{uav}}$ - an approach that adversely affects localization time and accuracy - we adopt a preprocessing step. Prior to localization, $\mathcal{I}_{\text{sat}}$ is divided into several local satellite maps $\mathcal{I}_{\text{cand}} = \{\mathcal{I}_{\text{sat}_1}, \mathcal{I}_{\text{sat}_2}, ..., \mathcal{I}_{\text{sat}_N}\}$, where $\mathcal{I}_{\text{sat}_i}$ denotes the $i$-th cropped local map and $N$ represents the total number of local maps. Each local map $\mathcal{I}_{\text{sat}_i}$ shares the same dimensions as $\mathcal{I}_{\text{uav}}$. To preserve salient features and avoid information loss during cropping, this study adopts a sliding window cropping approach with a duplication rate $r$. In our experiments, $r$ is set to 0.5.

\subsubsection{SuperPoint - Local feature extraction}
During the matching process, the absolute localization module employs SuperPoint~\cite{b12} as the local feature extractor to extract feature points $p$ and descriptors $d$ from the candidate local satellite map $\mathcal{I}_{\text{sat}_c}$ and the UAV image $\mathcal{I}_{\text{uav}}$. Specifically, the SuperPoint network utilizes a VGG13 architecture to initially encode the input image, producing the local feature representation. Subsequently, this local feature is fed into both the feature point decoder and the descriptor decoder. The feature point decoder, consisting of a convolutional layer followed by a softmax layer, generates the feature points $p$. Concurrently, the local feature is processed by the descriptor decoder, comprising a convolutional layer, an interpolation operation, and a normalization layer, to produce the descriptor $d$.
\begin{comment}
\begin{figure}[htp]
    \centering
    \includegraphics[width=\linewidth]{figs/SuperPoint.pdf}
    \caption{SuperPoint extracts the local features of input images.}
    \label{fig:superpoint}
\end{figure}
\end{comment}

\subsubsection{LightGlue - Local feature matching}
\begin{figure}[htp]
    \centering
    \includegraphics[width=\linewidth]{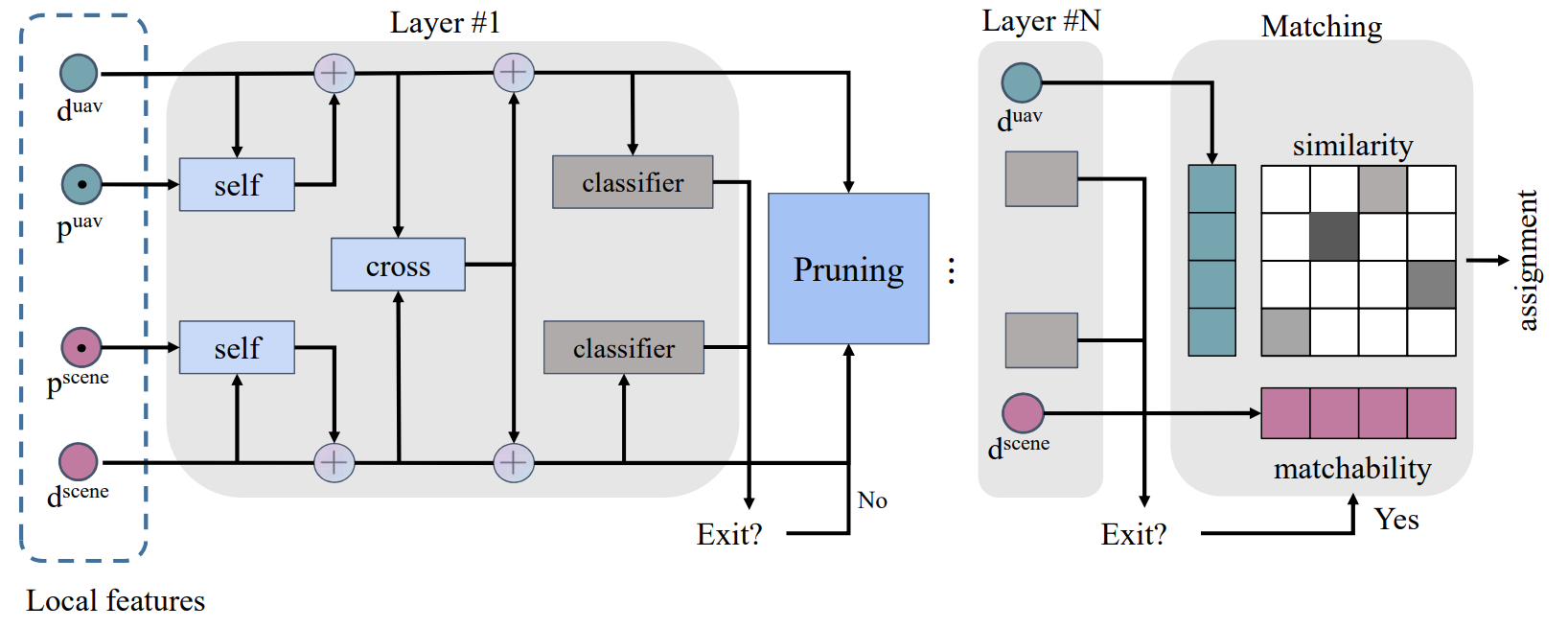}
    \caption{LightGlue matches the feature points.}
    \label{fig:lightglue}
\end{figure}

After extracting the feature points and descriptors from $\mathcal{I}_{\text{uav}}$ and $\mathcal{I}_{\text{sat}_c}$, the subsequent task involves feature matching to establish correspondences between feature points and estimate the homography matrix, which represents the transformation between $\mathcal{I}_{\text{uav}}$ and $\mathcal{I}_{\text{sat}_c}$. In this regard, this paper utilizes the LightGlue network~\cite{b11} for matching the UAV image with the local satellite map, as illustrated in Fig.~\ref{fig:lightglue}. The objective is to derive local matching degree $\sigma_c = [\sigma_1, \sigma_2, ..., \sigma_M]$ in the association and distribution of feature point pairs. $\sigma_i$ denotes the matching degree of the $i$-th pair of feature points and $M$ is the total number of the pairs of the feature points.

Upon the termination of the matching procedure, an assignment matrix and the matchability between the two sets of pruned feature points $\hat{p}^{uav}$ and $\hat{p}^{sat}$ are computed. The estimated homography matrix $\hat{H}$ between the UAV image and the local satellite map is then obtained via OpenCV~\cite{detone2016deep}, according to the feature points. Then the coordinate of the UAV $(P_x^G, P_y^G)$ in the global map can be estimated as
\begin{equation}
    \begin{aligned}
        \begin{pmatrix}
            P_x^G \\ 
            p_y^G \\
            1
        \end{pmatrix}^T
        =
        \begin{pmatrix}
            x_c \\
            y_c \\
            1
        \end{pmatrix}^T
        \times
        \hat{H} 
        +
        \begin{pmatrix}
            x_0 \\
            y_0 \\
            1
        \end{pmatrix}^T
    \end{aligned}
\end{equation}
where $(x_c, y_c)$ and the $(x_0, y_0)$ are the coordinates of the center point and upper left corner of $\mathcal{I}_{\text{sat}_c}$, respectively. The altitude $P_z^G$ of the UAV is obtained through the altimeter. To simulate real flight conditions, this paper introduces Gaussian error on the ground truth z-axis value to represent the altitude of the UAV. Therefore, the 3-DoF pose of the UAV estimated by absolute localization is
\begin{equation}
    \begin{aligned}
        P^G = (P_x^G, P_y^G, P_z^G + \frac{2}{\sqrt{\pi}} \int_0^z e^{-t^2} dt)
    \end{aligned}
\end{equation}
where the $\frac{2}{\sqrt{\pi}} \int_0^z e^{-t^2} dt$ is the Gaussian noise.

% 重点改
\subsubsection{Adaptive subsequent area prediction}\label{confidence}
\begin{figure}[htp]
    \centering
    \includegraphics[width=.8\linewidth]{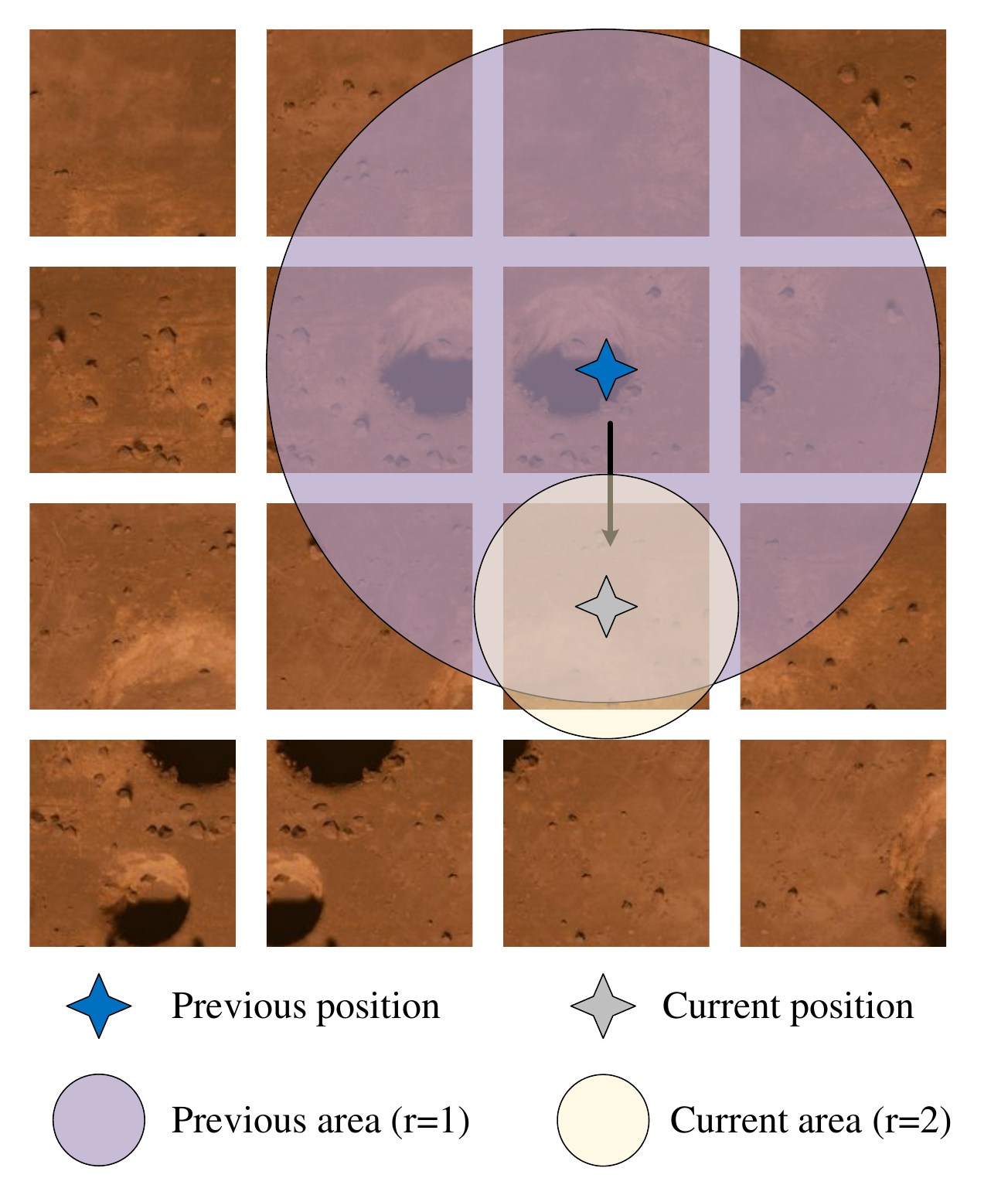}
    \caption{Schematic diagram of the subsequent area prediction strategy for large-scale image localization. The confidence level from the last UAV localization is 0.7. Accordingly, a search radius of $\lceil 10^{1-0.7}\rceil=2$ is employed, with the local map containing the current location serving as the center. This search aims to identify the local map that offers the closest match to the current UAV image. Subsequently, with the confidence of the current positioning standing at 0.9, the search radius is reduced to 1 when seeking the future UAV position.}
    \label{fig:pred}
\end{figure}

To expedite localization, we introduce an adaptive subsequent area prediction strategy tailored for large-scale image localization. Specifically, we devise an adaptive confidence for absolute localization results, which considers both global and local distribution characteristics among images. The local satellite map with the highest confidence is considered to be the location where the UAV is currently located. By utilizing the currently determined local satellite map as the center and defining the search range, generated based on confidence level $w$, as the radius, we predict the area of the UAV in the subsequent step, as shown in Fig.~\ref{fig:pred}.

The structural similarity (SSIM) between the shared regions of $\mathcal{I}_{\text{sat}_c}$ and $\mathcal{I}_{\text{UAV}}$ is utilized to gauge the similarity of the overall image distribution. Additionally, the average of the matching degrees of the feature points is employed to assess the similarity of the local feature points. Hence, the adaptive localization confidence is expressed as follows:
\begin{equation}
    \begin{aligned}
        w = SSIM(\mathcal{I}_{\text{sat}_c}, \hat{H}\mathcal{I}_{\text{uav}})\cdot \sum_{i=1}^M \sigma_i/M
    \end{aligned}
\end{equation}
Based on the confidence level $w$, the search radius $r\in [1, 10]$ of the potential area where the UAV may be located next is determined as follows:
\begin{equation}
    \begin{aligned}
        r = \lceil 10^{1-w} \rceil
    \end{aligned}
\end{equation}

\subsection{Absolute and relative poses fusion}\label{fusion}
The relative localization module employs ORB-SLAM2 with UAV images as input to estimate the local 6-DoF poses of the UAV. Upon receiving a UAV image, it is simultaneously processed by both the absolute localization module and the visual odometry tracking module. The outcomes from absolute localization, in conjunction with the pose estimations from visual odometry, are integrated into the LBA for back-end optimization. Through this process, rotation and translation matrices from the SLAM coordinate system to the world coordinate system are estimated using the trajectories obtained from absolute localization and visual odometry. This culminates in the derivation of a comprehensive UAV motion trajectory within the world coordinate system.

\subsubsection{Visual odometry-based feature tracking}
Visual odometry (VO) entails estimating UAV motion from UAV camera images by comparing consecutive image pairs. Meanwhile, ORB-SLAM2 leverages ORB features for real-time tracking. In VO, UAV pose estimation is relative to the initial pose within the SLAM coordinate system. However, optimizing the pose for each frame individually in the backend can considerably slow down the entire localization system. Therefore, only keyframes are forwarded to the backend local mapping thread. To ensure the adaptive selection of keyframes that accommodates both inter-frame matching in SLAM and absolute localization, this paper introduces an enhanced keyframe selection strategy based on the existing keyframe selection approach.

\begin{itemize}
    \item More than 10 frames have passed since the last keyframe
    \item The information entropy of the image satisfies 
    \begin{equation}
        H = -\sum_{i=0}^{255} \sum_{j=0}^{255} P_{i,j} / logP_{i,j} > 0.5
    \end{equation}
    where $P_{i,j} = \frac{f(i,j)}{W \times H}$ denotes the probability distribution of pixels
    \item The current keyframe has more than 90\% more points than the reference keyframe
\end{itemize}

\begin{comment}
\subsubsection{Trajectory alignment}
According to the obtained global positioning trajectory $q_{1,2...,n}$ and visual odometry trajectory $p_{1,2,...,n}$ where $n$ denotes the number of poses in each trajectory, the transformation matrix from the SLAM coordinate system to the world coordinate system can be estimated. Specifically, the purpose is to estimate the rotation matrix $R$ and the offset $t$ such that the objective function is optimal:
\begin{equation}
    \begin{aligned}
        (R, t) = argmin \frac{1}{n} \sum^n_{i=1}\begin{Vmatrix}q_i - (Rp_i + t)\end{Vmatrix}^2_2
    \end{aligned}
\end{equation}
where $q_i$ and $p_i$ denote the $i-th$ keyframe pose in global trajectory and visual odometry trajectory, respectively. It is noteworthy that our method maintains two sets of trajectories as a continuously updated global variable. This ensures that the transformation matrix from the SLAM coordinate system to the world coordinate system can be dynamically updated based on the most recent pose relationships. Using the estimated rotation and translation matrices, all poses estimated by the visual odometry can be transformed into the world coordinate system.
\end{comment}

\subsubsection{Pose fusion-based LBA}
LBA serves as an optimization technique employed to refine the accuracy of estimated camera poses and scene structures within the SLAM system. Its fundamental aim is to enhance the estimates of camera poses and map points by minimizing reprojection errors, thereby enhancing the precision and coherence of the SLAM system. The conventional BA algorithm optimizes the observed camera poses by constraining the disparity between the coordinates of the observed feature points and their corresponding reprojected coordinates:
\begin{equation}
    \begin{aligned}
        e_{i,j} = z_{i,j} - h(T_i, M_j)
    \end{aligned}
\end{equation}
where $e_{i,j}$ denotes the reprojection error of the 3D map point $M_j$ with the UAV camera extrinsic parameter $T_i$ corresponding to the $i-th$ image. $z \triangleq \left[ u_s, v_s \right]^T$ denotes the observed pixel coordinates of the feature point. $h(T, p)$ is the reprojection function, which reprojects the 3D point $P$ corresponding to the feature points into the camera plane according to the external matrix $T$ of the UAV camera.

To mitigate trajectory drift in the tracking module of monocular ORB-SLAM2 using the outcomes of absolute localization, this study adopts a pose fusion-based BA algorithm capable of integrating absolute coordinates. Based on the acquired absolute localization trajectory $q_{1,2...,m}$ and visual odometry trajectory $p_{1,2,...,m}$, where $m$ represents the total number of poses in each trajectory, the rotation and translation matrices from the SLAM coordinate system to the world coordinate system can be estimated. Specifically, the objective is to estimate the optimal rotation matrix $R$ and offset $t$ such that the objective function is optimized:
\begin{equation}
    \begin{aligned}
        (R, t) = argmin \frac{1}{n} \sum^T_{i=1}\begin{Vmatrix}q_{i} - (Rp_{i} + t)\end{Vmatrix}^2_2
    \end{aligned}
\end{equation}
where $q_{i}$ and $p_{i}$ denote the $i$-th keyframe poses in global trajectory and visual odometry trajectory, respectively. Our method maintains two sets of trajectories as continuously updated global variables. This ensures that the transformation matrix from the SLAM coordinate system to the world coordinate system can be dynamically adjusted based on the latest pose relationships. By utilizing the estimated rotation and translation matrices, all poses estimated by the visual odometry can be transformed into the world coordinate system.

Consequently, the cost function $f(x)$ to be optimized in the BA algorithm is formulated as follows:
\begin{equation}
    \begin{aligned}
        f(x) = \frac{1}{2}\sum^N_{i=1}\sum^M_{j=1}(e_{i,j}^Tw_ie_{i,j} + e_i^{G^T}w_{i}^Ge_{i}^G)
    \end{aligned}
\end{equation}
where $w_i$ is the confidence of observation $z_i$. In contrast to the traditional LBA algorithm, the improved LBA introduces global coordinate constraints as an additional factor in the optimization process by incorporating the deviation $e_i^G = \left[ e_x, e_y, e_z \right]$ with the absolute localization confidence $w_i^G$ between the results of the absolute localization and the visual odometry that is defined as $e_i^G = |P^G - P^L|$, where $P^G$ and $P^L$ are the results of absolute localization and visual odometry, respectively.

It is worth noting that due to the influence of the reprojection error function, this cost function is not linear. Therefore, in the actual optimization process, $f(x)$ is first linearized through a Taylor expansion and then refined step by step using the Levenberg-Marquard method.

\section{Experiment results}

\subsection{Experiment platform and hyperparameter settings}
The experiment platform operates within a Linux-based environment, utilizing Ubuntu 20.04 as the operating system. The experiments are conducted on a research server that features an Intel Core i9-9900K processor clocked at 3.6 GHz, 64GB of DDR4 RAM, and an NVIDIA GeForce RTX 2080Ti GPU.

\subsection{Dataset}
\begin{figure}[htp]
    \centering
    \includegraphics[width=.8\linewidth]{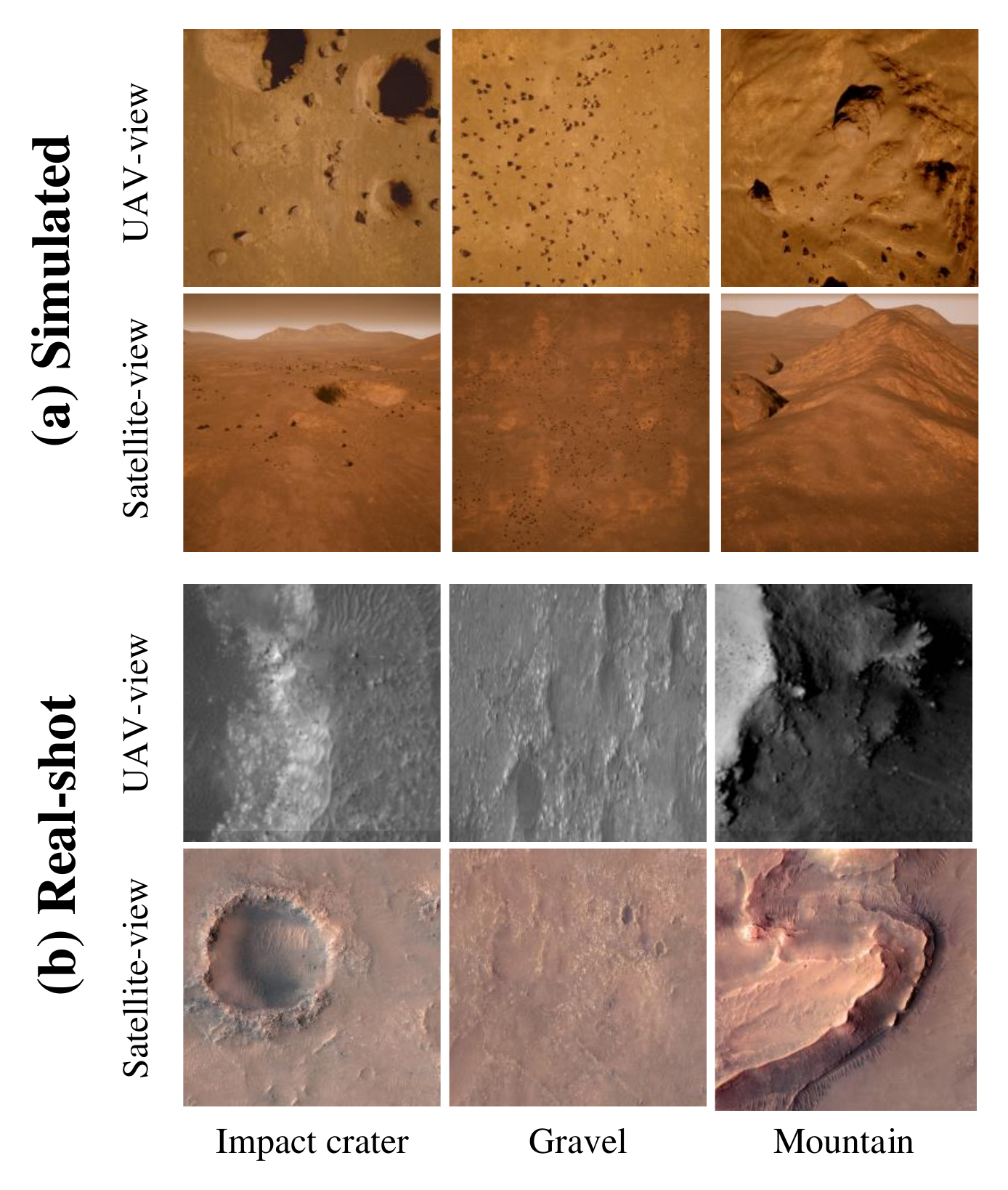}
    \caption{Samples of the proposed planetary UAV localization datasets. The first two rows are simulated satellite maps and UAV images, and the latter two rows are Martian surface images taken by the Perseverance lander\protect\footnotemark.}
    \label{fig:terrains}
\end{figure}

This paper produced a dataset including simulation data and real-shot data on Mars, some samples of which are shown in Fig.~\ref{fig:terrains}. We simulated three typical Martian terrains based on Unreal Engine 4 and Airsim, namely impact craters, gravel, and mountains. The UAV recorded overhead image sequences over these three terrains, including 925, 760, and 773 images respectively, with an average trajectory length of 149.63 meters. Our real-shot Mars data comes from images of the Martian surface taken by the lander of the Perseverance, including impact craters, gravels, and mountains. In each terrain, we selected 35 representative lander images and manually calibrated the homography matrix between them and the local map of Mars.

\footnotetext{https://mars.nasa.gov/mars2020/multimedia/raw-images/}

\subsection{Evaluation metrics}
\subsubsection{Metrics for image matching-based absolute localization}
For comparing image matching algorithms, we assess precision in absolute localization using homography estimation. We record error proportions for each method within 5 pixels (@5px), 10 pixels (@10px), 50 pixels (@50px) and 100 pixels (@100px). Furthermore, to assess the applicability of a localization algorithm in real-time systems, we also measured the time required for each algorithm to conduct a single matching operation.

\subsubsection{Metrics for visual localization}
To comprehensively compare visual localization algorithms, we use evo\cite{b2} for accuracy evaluation. By computing differences between estimated and ground truth trajectories, we derive statistics for five key metrics: standard deviation (std), root mean square error (rmse), minimum error (min), maximum error (max), and median error (median). These metrics offer a thorough assessment of various visual localization algorithm performances.

\subsection{Ablation study}

\subsubsection{Ablation study on pose fusion frequency}
To investigate the optimal fusion frequency for absolute and relative localization to ensure the safety of planetary exploration while saving energy as much as possible, we set the intervals for absolute localization to 1, 10, 50, and 100. The experimental results are shown in Table~\ref{tab:ablation_frequency}. 

\begin{table}[htp]
    \centering
    \caption{The results of ablation study on the pose fusion frequency. - indicates that positioning failed at this frequency.}
    \begin{tabular}{cccccc}
    \toprule
    Frequency & Max(m) & Rmse(m) & Std(m) & Mean(m) & Speed(Hz) \\ \hline
    1 & 1.003 & 0.219 & 0.133 & 0.184 &  17.887\\
    10 & 0.947 & 0.237 & 0.113 & 0.209 & 20.995 \\
    50 & 44.374 & 29.303 & 10.158 & 27.486 &  37.280\\
    100 & - & - & - & - & 62.813 \\
    \bottomrule
    \end{tabular}
    
    \label{tab:ablation_frequency}
\end{table}

It can be found that the positioning error obtained with a pose fusion frequency of 50 is too large, and SLAM failed to be successfully initialized in the experiment with a fusion frequency of 100. Therefore, this paper chooses 10 as the fusion frequency, which can accurately estimate the absolute 6-DoF pose at a speed of 20.995 Hz.

\subsubsection{Ablation study on adaptive area prediction}
To confirm the increased absolute localization speed from the area prediction strategy, we performed ablation experiments targeting the search range. Detailed results are in Table~\ref{tab:ablation_feedback}. The strategy significantly improved speed without significant accuracy loss, suggesting its suitability for real-time systems, enhancing speed while maintaining accuracy.

\begin{table}[ht]
    \centering
    \caption{The results of ablation study on the area prediction. $w/$ and $w/o$ denote the absolute localization methods with and without area searching, respectively.}
    \begin{tabular}{cccc}
    \toprule
         Terrain & Method & @1px / @5px / @10px & Time(ms) \\ \hline
         \multirow{2}{*}{Crater} & w/ & 0.2 / 0.5/ 0.9 &  81.91 \\
        \multirow{2}{*}{} & w/o & 0.2 / 0.5 / 0.9 & 793.12 \\  \hline

        \multirow{2}{*}{Gravel} & w/ & 0.2 / 0.4/ 1.0 &  90.41 \\
        \multirow{2}{*}{} & w/o & 0.3 / 0.6 / 1.0 & 868.11 \\  \hline

        \multirow{2}{*}{Mountain} & w/ & 0.0 / 0.5/ 0.8 &  81.91 \\
        \multirow{2}{*}{} & w/o & 0.2 / 0.7 / 1.0 & 770.92 \\ 
    \bottomrule
    \end{tabular}
    \label{tab:ablation_feedback}
\end{table}

\subsubsection{Ablation study on absolute localization confidence}\label{fusion-frequency}
In order to explore the impact of the proposed confidence mechanism on positioning accuracy, we compare the proposed confidence with 5 fixed confidences, SSIM-based confidence, and matching degree-based confidence. The results of this comparative analysis are presented in Table.~\ref{tab:ablation_confidence}. The approach using absolute localization confidence as LBA reference frame weighting outperforms methods with fixed confidence.

\begin{table}[htp]
    \centering
    \caption{The results of the ablation study on the absolute localization confidence. $Adaptive$ denotes our method that employs the adaptive absolute localization confidence.}
    \begin{tabular}{ccccc}
    \toprule
    Confidence & Max(m) & Rmse(m) & Std(m) & Mean(m) \\ \hline
    0.2 & 8.260 & 3.784 & 1.872 & 3.281 \\
    0.4 & 2.898 & 0.798 & 0.462 & 0.643 \\
    0.6 & 4.344 & 1.332 & 0.675 & 1.139 \\
    0.8 & 4.631 & 2.278 & 1.037 & 2.016 \\
    1.0 & 3.655 & 1.084 & 0.581 & 0.906 \\
	SSIM & 0.948 & 0.256 & 0.124 & 0.225 \\
	Matching degree & \textbf{0.896} & {0.266} & {0.122} & {0.235} \\
    Adaptive & {0.947} & \textbf{0.237} & \textbf{0.113} & \textbf{0.209} \\
    \bottomrule
    \end{tabular}
    \label{tab:ablation_confidence}
\end{table}

\subsection{Comparison experiments}

\subsubsection{Image matching-based absolute localization experiment}
The absolute localization module relies heavily on matching UAV and satellite images, making the choice of a robust image-matching method crucial. We select 30 pairs of simulated UAV-local satellite images from different terrains and 20 pairs of real-shot ones from the lander of the Ingenuity as test samples.

We opt for representative image matching methods, encompassing algorithms founded on traditional operators like Scale Invariant Feature Transform (SIFT)~\cite{b4}, sparse feature-based methods like D2Net~\cite{b5}, R2D2~\cite{b6}, SuperGlue~\cite{b7}, and LightGlue~\cite{b11}, and semi-dense matching methods like DFM~\cite{b8}, LoFTR~\cite{b9}, and COTR~\cite{b10}. The experimental results of different image matching methods in various terrains are shown in Table~\ref{tab:immatch}. Evidently, SIFT, SP+SG, and SP+LG outperform other methods in terms of matching accuracy. Among these three methods, SP+LG has the fastest calculation speed, which is more obvious in the real-shot scene. Therefore, this paper adopts the SP+LG approach as the backbone of the absolute localization module.

\begin{table}[htp]
    \centering
    \caption{Experimental results of different image matching methods on simulated and real-shot scenes, whereas \textit{@X} means the proportion of matching error in \textit{X} pixels. The SP, SG, and LG denote the SuperPoint, SuperGlue, and LightGlue, respectively.}
    \begin{tabular}{ccccc}
    \toprule
         \multicolumn{1}{c}{\multirow{2}{*}{Method}} &
         \multicolumn{2}{c}{Simulation} &
         \multicolumn{2}{c}{Mars} \\ \cline{2-3} \cline{4-5}

         \multirow{2}{*}{} 
         & @5 /@10 & Time(ms)
         & @10 /@50 /@100 & Time(ms) \\ \hline

        SIFT + nn  & 0.87 /0.97 & 90.47 & 0.30 /0.80 /0.80 & 737.12\\
        D2Net & 0.70 /0.83 & 270.01 & 0.00 /0.20 /0.20 & 2983.52\\
        R2D2 & 0.60 /0.63 & 390.71 & 0.00 /0.00 /0.00 & - \\ 
        DFM & 0.83 /0.93 & 122.44 & 0.13 /0.38 /0.38 & 604.34\\
        LoFTR & 0.60 /0.63 & 191.13 & 0.00 /0.00 /0.00 & -\\
        SP+SG & 0.93 /0.97 & 183.19 & 0.10 /0.60 /0.60 & 328.60\\
        SP+LG & 0.90 /0.97 & \textbf{74.87} & 0.20 /0.60 /0.80 & \textbf{269.95}\\
    \bottomrule
    \end{tabular}
    \label{tab:immatch}
\end{table}

\subsubsection{Visual localization experiment}
We compare JointLoc against SLAM-based ORB-SLAM2\cite{b14} and ORB-SLAM3\cite{b15}, and image matching-based SIFT, SuperGlue, LoFTR, and LightGlue across 3 different terrains. We compared these algorithms from the perspective of positioning accuracy, such as mean (m), rmse (m) and std (m), and positioning speed (Hz). The experimental results can be seen in Table.\ref{tab:localization}.

\begin{table}[htp]
\caption{The average of position accuracy (m) and speed (Hz) of different localization methods in 3 simulated terrains.}
    % \centering
    \renewcommand\arraystretch{1.5}
    \begin{tabular}{cccc}
    \toprule
        Category & Method & Mean / Rmse / Std & Speed \\ \hline

        \multirow{2}{*}{Monocular SLAM} & ORB-SLAM2 
            & 0.485 /0.594 /0.342 & 22.191 \\
        \multirow{2}{*}{} & ORB-SLAM3 
            & 0.420 /0.557 /0.361 & \textbf{30.380}\\ \hline
        
        \multirow{4}{*}{Absolute localization} & SIFT 
            & 5.776 /6.532 /3.051 & 1.357 \\
        
        \multirow{4}{*}{} & SuperGlue 
            & 5.289 /6.024 /2.855 & 3.043\\
        
        \multirow{4}{*}{} & LoFTR 
            & - & -\\
        \multirow{4}{*}{} & LightGlue 
            & 5.562 /6.253 /2.856 & 3.704\\ \hline

        \multirow{1}{*}{Fusion localization} & JointLoc 
            & \textbf{0.209} /\textbf{0.237} /\textbf{0.113} & 20.995 \\

    \bottomrule
    \end{tabular}
    \label{tab:localization}
\end{table}

The experimental results reveal that JointLoc surpasses both ORB-SLAM2 and ORB-SLAM3 on most metrics. Meanwhile, LoFTR relying on depth features, encounters localization failures in textureless regions due to its dependence on feature extraction. Moreover, the entirely matching-based approach of SIFT, SuperGlue, and LightGlue exhibits inadequate processing speed during testing, rendering real-time performance unattainable, and their localization errors are notably higher. In contrast, JointLoc maintains a processing speed of 20.995 Hz, accurately estimating the 6-DoF pose.

\subsection{Timimg results}
We have also counted the average running time of each module of JointLoc across various scenarios, as depicted in Table~\ref{tab:timecost}.

\begin{table}[htp]
\caption{The average running time (ms) of each module of JointLoc in 3 simulated terrains. AbsLoc and RelLoc denotes the absolute localization module and the relative localization module, respectively.}
    % \centering
    \renewcommand\arraystretch{1.5}
    \begin{tabular}{ccccc}
    \toprule
        Module & Uint & Crater / Gravel / Mountains \\ \hline

        \multirow{2}{*}{AbsLoc} & Feature extraction 
            & 18.922 / 14.768 / 12.505 \\
        \multirow{2}{*}{} & Feature matching 
            & 42.773 / 29.762 / 37.665 \\ \hline
        
        \multirow{4}{*}{RelLoc} & Tracking 
            & 28.599 / 29.364 / 26.910 \\
        
        \multirow{4}{*}{} & Mapping 
            & 212.550 / 211.579 / 210.452\\
        
        \multirow{4}{*}{} & Looping 
            & 207.995 / 212.817 / 223.534 \\
        \multirow{4}{*}{} & Pose fusion-based LBA 
            & 403.045 / 398.162 / 412.116 \\
    \bottomrule
    \end{tabular}
    \label{tab:timecost}
\end{table}
\section{CONCLUSIONS}
This paper introduces JointLoc, a robust and dependable localization framework used for UAVs to cope with the absence of the GNSS in planetary UAV navigation. JointLoc stands out for its integration of both relative and absolute localization modules, showcased within this paper, while also offering compatibility with a wide array of existing localization techniques for effortless module replacement. Through the adaptive fusion of relative and absolute localization modules, JointLoc enables real-time estimation of the 6-DoF pose of the UAV in the planetary coordinate system. JointLoc achieves an average positioning error of 0.237 meters at 21 Hz, better than ORB-SLAM2 (0.594 m) and ORB-SLAM3 (0.557 m). Moreover, a specialized UAV dataset within a planetary context is proposed, including both simulation data from AirSim and the images of the surface of Mars captured by the lander of the Perseverance, further enriching related research fields.

\section{Acknowledgements}
Thanks to the Mars Exploration Program of NASA, the real-shot part of our dataset is based on the images captured by the Ingenuity helicopter.

\end{document}